\documentclass[sigconf]{acmart}

\copyrightyear{2021}
\acmYear{2021}
\setcopyright{acmcopyright}
\acmConference[CIKM '21] {Proceedings of the 30th ACM International Conference on Information and Knowledge Management}{November 1--5, 2021}{Virtual Event, Australia.}
\acmBooktitle{Proceedings of the 30th ACM Int'l Conf. on Information and Knowledge Management (CIKM '21), November 1--5, 2021, Virtual Event, Australia}
\acmPrice{15.00}
\acmDOI{10.1145/3459637.3482126}
\acmISBN{978-1-4503-8446-9/21/11}

\usepackage{multicol,lipsum}
\usepackage{graphicx}
\usepackage{subfigure}
\usepackage{booktabs} 
\usepackage{latexsym}
\usepackage{makecell} 
\usepackage{amsfonts} 
\usepackage{amsmath} 
\usepackage{hyperref}
\usepackage{bm}
\usepackage{tabularx}
\usepackage[capitalise]{cleveref}
\usepackage{paralist}
\usepackage[super]{nth}
\usepackage{wrapfig}

\DeclareMathAlphabet{\pazocal}{OMS}{zplm}{m}{n}

\setcopyright{none}

\usepackage{microtype}
\usepackage{graphicx}
\usepackage{subfigure}
\usepackage{booktabs} 

\usepackage{amsmath} 
\usepackage{amsfonts}
\usepackage{bm}
\usepackage{multirow} 
\usepackage{hyperref}

\def\xx{Grad-SAM}
\newcommand{\R}{\mathbb{R}}
\newcommand{\T}{\mathcal{T}}

\usepackage{latexsym}
\usepackage{amsmath}
\usepackage{amsfonts}
\usepackage{multirow}
\usepackage[shortlabels]{enumitem}
\usepackage{verbatim}
\usepackage{latexsym}
\usepackage{amsmath}
\usepackage{dsfont}
\usepackage{graphicx}
\usepackage{booktabs}
\usepackage{dirtytalk}
\usepackage{amsmath, graphicx}
\usepackage{tikz}
\usepackage{url}
\usepackage{xspace}

\usepackage{xcolor}
\usepackage{makecell}
\usepackage{algorithm}
\usepackage[noend]{algpseudocode}
\newcounter{algsubstate}


\usepackage[utf8]{inputenc}

\begin{document} 
\fancyhead{}
\title{Grad-SAM: Explaining Transformers via \\ Gradient Self-Attention Maps}

\author{Oren Barkan}
\authornote{Authors contributed equally to this research.}
\affiliation{%
  \institution{The Open University}
  \country{Israel}
}

\author{Edan Hauon}
\authornotemark[1]
\affiliation{%
  \institution{IDC Herzeliya}
  \country{Israel}
}

\author{Avi Caciularu}
\authornotemark[1]
\affiliation{%
  \institution{Bar-Ilan University}
  \country{Israel}
}

\author{Ori Katz}
\affiliation{%
  \institution{Microsoft \& Technion}
  \country{Israel}
}

\author{Itzik Malkiel}
\affiliation{%
  \institution{Tel Aviv University}
  \country{Israel}
}

\author{Omri Armstrong}
\affiliation{%
  \institution{Tel Aviv University}
  \country{Israel}
}

\author{Noam Koenigstein}
\affiliation{%
  \institution{Microsoft \& Tel Aviv University}
  \country{Israel}
}








\renewcommand{\shortauthors}{Barkan and Haun, et al.}
 \fancyhead{} 
\setcopyright{acmcopyright}

\begin{abstract}
  Transformer-based language models significantly advanced the state-of-the-art in many linguistic tasks. As this revolution continues, the ability to explain model predictions has become a major area of interest for the NLP community. In this work, we present Gradient Self-Attention Maps (Grad-SAM) - a novel gradient-based method that analyzes self-attention units and identifies the input elements that explain the model's prediction the best. Extensive evaluations on various benchmarks show that Grad-SAM obtains significant improvements over state-of-the-art alternatives.
\end{abstract}



\maketitle

\section{Introduction}
Deep contextualized language models have significantly advanced the state-of-the-art in various linguistic tasks such as question answering \citep{yang2019xlnet}, coreference resolution \citep{joshi-etal-2019-bert}, and other NLP benchmarks~\cite{wang-etal-2018-glue,wang2019superglue}. These models provide an efficient way to learn representations in a fully self-supervised manner from text corpora, solely using co-occurrence statistics. 
Empirically, deep contextualized language models that rely on the Transformer architecture~\cite{vaswani2017attention}, were shown to achieve state-of-the-art results, when are finetuned on supervised tasks~\cite{NIPS2015_7137debd,peters-etal-2018-deep,radfordimproving,barkan2020cold,barkan2020scalable}.  

Unlike traditional feature-based machine learning models that assign and optimize weights to interpretable explicit features, Transformer based architectures such as BERT~\cite{devlin-etal-2019-bert} rely on a stack of multi-head self-attention layers, composed of hundreds of millions of parameters. These models are much more complex and computational heavier than models that learn non-contextualized representations~\cite{mikolov2013distributed,pennington2014glove,barkan2017bayesian,barkan-etal-2020-within,barkan-etal-2020-bayesian,vbn,acf,cwh,barkan2021cold}. 

At inference, Transformers-based models compute pairwise interactions of the resulting vector representations, making it particularly challenging to explain which part of the input contributed to the final prediction.
Recently, significant efforts were put towards interpreting these models, mostly by applying \emph{white-box} analysis~\cite{Wallace2019AllenNLP,ebrahimi2018hotflip}. In this case, the goal is to probe the models' performance through lower-level components in the neural model.

Nonetheless, a central line of works attempted to study the types of linguistic knowledge encoded in such deep language models. Recent studies discovered that the BERT model~\cite{devlin2019bert} was shown to rely on surface structures (word, order, specific sequences, or co-occurrences) during pre-training~\cite{rogers2020primer,clark2019what}. However, how and where exactly this information is stored, as well as retrieved during inference time, is still an open question that yet to be explored.

Gradient-based methods~\cite{sundararajan} yield decent ad-hoc explanations for predictions by highlighting which parts of the input correspond with the model's predictions. Moreover, recent works~\cite{abnar-zuidema-2020-quantifying,wiegreffe-pinter-2019-attention,serrano-smith-2019-attention} showed that the input's sequence gradients have a high correlation with importance scores provided by human annotators, providing better interpretations than the scores produced by the raw token attentions. They observed that gradient-based ranking of attention scores better explains the model prediction than their magnitudes.

In this paper, we propose Gradient Self-Attention Maps (\xx) - a novel gradient-based interpretation method that probes BERT's predictions. We demonstrate the effectiveness of Grad-SAM as a ranking machinery that identifies the input elements that contribute to the model prediction the most. We present both quantitative and qualitative results, indicating that Grad-SAM significantly outperforms state-of-the-art alternatives.

\section{Related Work}

Recent methods for explaining predictions made by deep learning models considered explanations computed through convolutional layers ~\cite{selvaraju2017grad,patro2019u,patro2020explanation,nam,gam}, and attention based architectures~\cite{pruthi2019learning,wiegreffe-pinter-2019-attention,tai2020exbert,clark2019what,htut2019attention,Wallace2019AllenNLP,ebrahimi2018hotflip,tai2020exbert, barkan2020scalable, barkan2020explainable, barkan2020attentive,malkiel2020optimizing,ginzburg2021self}.
While the authors in \cite{jain2019attention} argue that attention scores sometimes does not interpret model predictions faithfully, other works show that attention scores do offer plausible and meaningful interpretations that are often sufficient and correct \cite{pruthi2019learning,wiegreffe-pinter-2019-attention,vig2019visualizing}.

Recently, a framework attempting to rationalize predictions named FRESH~\cite{jain-etal-2020-learning} was proposed. FRESH equips BERT predictions with faithfulness by construction -- their goal was to focus on extracting rationales by introducing an additional extractor model trained to predict snippets. Then, a classifier is trained over the snippets and is expected to output faithful explanations.

In parallel, several methods were suggested to produce model interpretations via gradients~\cite{Wallace2019AllenNLP}. The vast majority of these works utilized the gradients of the prediction w.r.t. the input for computing the importance of each token in the input sequence. In some scenarios, gradient-based approaches were shown to provide more faithful explanations than attention-based methods~\cite{bastings-filippova-2020-elephant}. This family of gradient-based explainability methods have been applied \cite{li2021normal,chan2021salkg,chrysostomou2021variable}, yet in a task-specific manner, to different downstream tasks.

Unlike the aforementioned works, our proposed Grad-SAM method integrates the information from attention scores together with their gradients in a finetuned BERT model. Furthermore, Grad-SAM is generic (not task-specific) in the sense it relies on the analysis of a given finetuned model only, and does not require the training an additional extractor network (in contrast to FRESH). Yet, our evaluation shows that \xx \xspace provides more faithful rationales than the ones produced by FRESH, across multiple linguistic tasks. 
Finally, it is worth noting that Grad-SAM, in its essence and purpose, differs from ~\cite{selvaraju2017grad} by several aspects: First, it focuses on the NLP domain. Second, it operates on a completely different architecture (BERT). Third, it analyzes self-attention units from multiple layers in the model and not just the last one. Lastly, Grad-SAM treats negative gradient differently (Sec.~\ref{sec:gar}).

\section{Gradient Self-Attention Maps}
\label{sec:gar}
We begin by defining the problem setup and several notations. Then, we describe and explain \xx\xspace in detail. While our focus is on providing explanations for BERT, \xx\xspace is applicable for any architecture based on self-attention (SA) units \cite{vaswani2017attention}.

Let $\mathcal{T}=\{t_i\}_{i=1}^{N_{\T}}$ be a vocabulary of supported tokens. Let $\Omega$ be a set containing all sentences of length $N$ that can be composed from $\mathcal{T}$ (shorter sentences are padded by a reserved token {\tt [PAD]}), where each sentence starts and ends with the special tokens {\tt [CLS]} and {\tt [SEP]}, respectively.
BERT~\cite{devlin2019bert} is a parametric function $s:\Omega\rightarrow\R^n$ that receives a sequence (sentence) of $N$ tokens $x=(x_i)_{i=1}^N$ ($x_i\in \T$) and outputs a $n$-dimensional vector of scores $s_x:=s(x)$. In general, BERT is optimized via a two-phase process: In the \emph{pre-training} phase, BERT is optimized w.r.t. to the Masked Language Model task together with the Next Sentence Prediction task. In the second phase, BERT is \emph{finetuned} w.r.t. a specific downstream task e.g., multiclass/binary classification, or a regression task. Hence, $n$ stands for the number of classes/output dimension
and changes w.r.t. the specific downstream task at hand. 

$s$ is implemented as a cascade of $L$ encoder layers. Given a sentence $x\in\Omega$, each token $x_i$ in the sentence is mapped to a $d$-dimensional vector (embedding) to form a matrix $U^0_x\in\R^{d \times N}$. In practice, this embedding is a  summation of the token, positional, and segment embeddings. Then, $U^0_x$ is passed through a stack of $L$ encoder layers. The $l$-th encoder layer ($1\leq l \leq L$) receives the intermediate  representations $U_x^{l-1}\in\R^{d \times N}$ (produced by the $(l-1)$-th layer), and outputs the new representations $U_x^l$. Finally, $u_{\text{{\tt [CLS]}}}^L$ (the first column in $U_x^L$, which corresponds to the {\tt [CLS]} token) is used as input to a subsequent fully connected layer that outputs $s_x$. 

Each encoder layer employs $M$ SA heads that are applied in parallel to $U_x^{l-1}$, producing $M$ different attention matrices
\begin{equation}
\label{eq:att-scores}
    A_x^{lm}=\text{softmax}\left(\frac{(W^{lm}_qU_x^{l-1})^TW^{lm}_kU_x^{l-1}}{\sqrt{d_a}}\right),
\end{equation}
where $W^{lm}_q,W^{lm}_k\in\R^{d_a \times d}$, are the query and key mappings, and $1\leq m \leq M$. Each entry $[A_x^{lm}]_{ij}$ quantifies the amount of attention $x_i$ receives by $x_j$, according to the attention head $m$ in the layer $l$. Then, the encoder output $U_x^{l}$ is obtained by a subsequent set of operations that involves the $M$ attention matrices and \emph{value} mappings as detailed in  \cite{vaswani2017attention}. We refer to ~\cite{devlin2019bert} for a detailed description of BERT.

Our goal is to explain the predictions made by BERT. To this end, we propose to utilize the attention matrices $A_x^{lm}$ together with their gradients in order to produce a ranking over the tokens in $x$ s.t. tokens that affect the model prediction the most, are ranked higher.

Given a sentence $x\in\Omega$ and a finetuned BERT model $s$, we compute the the prediction $s_x\in\R^n$. In this work, our focus is on classification tasks. Specifically, for binary classification, $n=1$, and we set $s_x$ as the logit score. However, for multiclass classification, $n>1$ (depending on the number of distinct classes) we focus on a specific entry in $s_x$ which is associated with the ground truth class to be explained. For the sake of brevity, from here onwards, $s_x$ represents the logit score in binary classification, or the logit score associated with the ground truth class $s_x$ (in the case of multiclass classification) and disambiguation should be clear from the context. 

Our goal is to quantify the importance of each token $x_i\in x$ w.r.t. $s$. In other words, we wish to identify tokens in $x$ that contribute to $s$ the most, hence explaining the prediction made by the model. To this end, we propose the following explanatory scheme: First, we pass $x$ through BERT to compute $s_x$. Then, the \emph{importance} of the token $x_i$ w.r.t. the prediction $s_x$ is computed by:
\begin{equation}
\label{eq:exp-score}
    r_{x_i}=\frac{1}{LMN}\sum_{l=1}^L \sum_{m=1}^M \sum_{j=1}^N [H_x^{lm}]_{ij},
\end{equation}
with
\begin{equation}
\label{eq:exp-map}
H_x^{lm}=A_x^{lm} \circ \text{ReLU}(G_x^{lm}),
\end{equation}
where $G_x^{lm}:=\frac{\partial s_x}{\partial A_x^{lm}}$ are the element-wise gradients of $s_x$ w.r.t. to $A_x^{lm}$, and $\circ$ stands for the Hadamard product.
Eq.~\ref{eq:exp-score} scores the importance of each token $x_i\in x$ w.r.t. $s_x$, enabling ranking the tokens in $x$ according to their importance. Higher values of $r_{x_i}$ indicate higher importance of $x_i$, hence a better explanation of the prediction score $s_x$. In practice, for $x_i\in\{\tt{[CLS],[SEP],[PAD]}\}$, we set $r_{x_i}=-\infty$, as these tokens cannot provide for good explanations.

The motivation behind Eqs.~\ref{eq:exp-score} and \ref{eq:exp-map} is as follows: We are willing to identify tokens in $x$ for which 1) High attention is received from other tokens in $x$ (information from the \emph{attention activations}), and 2) Further increase in the amount of the received attention will increase $s_x$ the most (information from \emph{gradient} of the attention activations).
Eq.~\ref{eq:exp-map} ensures that these two conditions are met, since if $[G_x^{lm}]_{ij}\leq0$, then $[H_x^{lm}]_{ij}=0$, and if $[A_x^{lm}]_{ij}$ is small, then $[H_x^{lm}]_{ij}$ is close to zero (recall that $[A_x^{lm}]_{ij}\geq0$, as it is the result of softmax). Finally, Eq.~\ref{eq:exp-score} aggregates the overall contribution of the attention scores and the positive gradients from all SA heads across all encoder layers, w.r.t. $x_i\in x$.

We wish to re-emphasize the following important point: Zeroing the negative gradients in Eq.~\ref{eq:exp-map} enables the preservation of the positive values of $H_x^{lm}$ (associated with positive gradients), which otherwise may be cancelled out by a large accumulated negative value in the summation in Eq.~\ref{eq:exp-score}. 
The activations in the $i$-th row within an attention matrix $A_x^{lm}$ quantify the importance of the token $x_i$ w.r.t. the other tokens in $x$. In addition, if $[G_x^{lm}]_{ij}>0$, then an increase in the activation $[A_x^{lm}]_{ij}$ should lead to an increase in the model’s output score. Therefore, the importance of the token $x_i$ (according to the attention head $m$ in the encoder layer $l$) is determined by the summation over the $i$-th row in $H_x^{lm}$, and the contribution to this sum come from elements for which both the activation and its gradient are positive. Finally, the overall importance of $x_i$ is accumulated from the $M$ SA heads in $L$ layers according to Eq.~\ref{eq:exp-score}.

In regular BERT-base models, there are 144 SA heads ($M=12,L=12$) that act as filters. However, in practice, we observed that only a few attention entries are activated. Specifically, we found out that there is a large number of activations that are close to zero, but associated with negative gradients. The accumulated effect of this negative sum leads to a suppression (or even complete cancellation) of the small number of activations with positive gradients (which hold the actual information we are wish to preserve). Hence, we zero those negative gradients (using ReLU). The necessity of the negative gradient trimming, prior to the summation, along with the complementary contribution from the attention activations and their gradients, are validated in the ablation study presented in Tabs.~\ref{table:main-results} and \ref{table:AOPC-results}.

\section{Experimental Setup and Results}
\label{sec:results}

\subsection{Datasets and Downstream Tasks}
\label{sec:datasets}
In all of the experiments, we use a pre-trained BERT-base-uncased model, taken from Huggingface's Transformers library~\cite{wolf-etal-2020-transformers}, associated with a standard tokenizer. Then, we finetune BERT on five downstream tasks (in separate):
\begin{itemize}[leftmargin=0pt,wide=0pt]
 \setlength\itemsep{-1pt}
\item The Stanford Sentiment Treebank \textbf{(SST)}~\cite{socher-etal-2013-recursive}: A sentiment analysis task (binary classification).
\item AgNews \textbf{(AGN)}~\cite{del2005ranking}: A multiclass classification task, where news articles are categorized into \emph{science, sports, business, world}.

\item \textbf{IMDB}~\cite{maas-EtAl:2011:ACL-HLT2011}: A sentiment analysis task (binary classification of movie reviews).
\item MultiRC \textbf{(MRC)}~\cite{khashabi-etal-2018-looking}: A binary classification task. The same processing from ~\cite{jain-etal-2020-learning} was followed to produce True / False labels w.r.t. a given snippet.
\end{itemize}
\subsection{Evaluated Methods}
\label{sec:eval_methods}
We compare several methods for ranking the importance of tokens in a sentence $x$:
\begin{enumerate}[leftmargin=0pt,wide=0pt]
 \setlength\itemsep{-1pt}
    \item \textbf{Gradient}: This is the 'Gradient' method from ~\cite{jain-etal-2020-learning}. 
    \item \textbf{{{\tt [CLS]}}} \textbf{Att}: This is the '{{\tt [CLS]}} Attn' method from \cite{jain-etal-2020-learning}.
    \item \textbf{Att}: Setting $H_x^{lm}=A_x^{lm}$ and using Eq.~\ref{eq:exp-score}. 
    \item \textbf{Att-Grad}: Setting $H_x^{lm}=G_x^{lm}$ and using Eq.~\ref{eq:exp-score}.
    \item \textbf{Att-Grad-R}: Setting $H_x^{lm}=\text{ReLU}(G_x^{lm})$ and using Eq.~\ref{eq:exp-score}.
    \item \textbf{Att $\times$ Att-Grad}: Setting $H_x^{lm}=A_x^{lm} \circ G_x^{lm}$ and using Eq.~\ref{eq:exp-score}. 
    \item \textbf{Grad-SAM}: Using Eq.~\ref{eq:exp-score} (our proposed method).
\end{enumerate}
Note that methods 3-6 are ablated versions of Grad-SAM.
\subsection{Quantitative Evaluations}
\label{subsec:setup_eval}

Our first evaluation follows the protocol from \cite{jain-etal-2020-learning}: For each sentence $x$ in the test set, we used each method (Sec.~\ref{sec:eval_methods}) to produce a different ranking over $x$'s tokens. Then, we preserved the top $k$\% ranked tokens and masked the rest. We used the same values from \cite{jain-etal-2020-learning}: $k=20\%$ for SST, AGN, and MRC, and $k=30\%$ for IMDB. Finally, we compute the mean Macro F1 score for each combination of method and task. We further include the original \emph{Full text} results obtained on $x$ without masking.

Table~\ref{table:main-results} depicts the results for each combination of explanation method and task. First, we see that \xx\xspace significantly outperforms both the Gradient and the {\tt {[CLS]}} Att methods from \cite{jain-etal-2020-learning}. It is worth noting that the methods from \cite{jain-etal-2020-learning} require a simultaneous training of an auxiliary model that \emph{learns to mask}, during the BERT's finetuning phase, while our Grad-SAM method eliminates this need.
The ablation study reveals that the \text{ReLU} operation over the gradient-attention is crucial (Att-Grad-R $\succ$ Att-Grad). This indicates that by trimming the negative gradients, we avoid the unwanted suppression of positive gradients (if exist) across the summation in Eq.~\ref{eq:exp-score}. Moreover, Grad-SAM, which combines the attention scores together with \text{ReLUed} gradients, performs the best across all tasks.

\begin{table}[t]
    \centering
    \begin{tabular}{lccccc}
        \toprule
        Method   &                SST  &            AGN &                         IMDB &           MRC \\
        \midrule
        \emph{Full text} &  .904  &  .942  &  .957 &   .682  \\
        \midrule
         Gradient&  .682  &  .863 &  .933   &   .654 \\
         {\tt [CLS]} Att & .812  &  .911  &  .941  & .639  \\
         \midrule
          Att &  .801 &  .855 &  .837  &  .632 \\
          Att-Grad &  .706  &  .792  &  .715  &  .634\\

         Att-Grad-R&  \underline{.819}  &  \underline{.911}  & \underline{.946}  &   \underline{.657}  \\
         Att$\times$Att-Grad& .810  &  .778 &  .743  &  .636\\

        \xx &  \textbf{.823 } &  \textbf{.921 } &  \textbf{.949 } &   \textbf{.662 } \\

        \bottomrule
    \end{tabular}
    \caption{Model predictive performances across datasets. We report the mean-macro F1 scores on the test sets. The top row (\emph{Full text}) corresponds to passing the sentence, without masking (upper-bound on performance).}
    \label{table:main-results}
\end{table}

\begin{table}[t]
    \centering
    \begin{tabular}{lccccc}
        \toprule
        Method   &                SST  &            AGN &                        IMDB &           MRC \\
        \midrule
        \emph{Full text} &  .904  &  .942 &    .957 &   .682  \\
        \midrule
         Gradient&  .16  &  .101 &   .05  &   .09 \\
         {\tt [CLS]} Att & .165  &  \textbf{.177}  &  .055 & .082  \\
         \midrule
          Att &  .113 &  .118 &  .047  &  .093 \\
          Att-Grad &  .072  &  .109   &  .031  &  .113\\

         Att-Grad-R&  \underline{.179}  &  .132  &   \underline{.059}  &  \underline{.12}  \\
         Att$\times$Att-Grad& .152  &  .12 &   .052  &  .103\\

        \xx &  \textbf{.195 } &  \underline{.14 } &    \textbf{.065 } &  \textbf{.122 } \\

        \bottomrule
    \end{tabular}
    \caption{AOPC evaluation. Note that the \emph{Full Text} row is presented for reference, reporting the mean-macro F1 scores on test sets without any word filtration. The other rows report the AOPC for each combination of method and dataset.} 
    \label{table:AOPC-results}
\end{table}
Our second evaluation is based on the Area Over the
Perturbation Curve (AOPC) \cite{nguyen-2018-comparing} metric that is designed to assess the faithfulness of explanations produced by Grad-SAM and the other methods. AOPC calculates the average change of accuracy over test data by masking the top k\% tokens in the sentence $x$ (the tokens are ranked by the explanation method). Hence, the larger the value of AOPC, the better the explanations of the models. Table \ref{table:AOPC-results} depicts the results with k = 20\% (the top 20\% of words ranked by each method). The \emph{Full Text } row is presented for reference, reporting the mean-macro F1 scores on the original sentences from the test sets without any masking. The other rows report the AOPC for each combination of method and dataset.
Again, we compare Grad-SAM to the same baselines from Sec.~\ref{sec:eval_methods} (and perform an ablation study).

The results in Tab.~\ref{table:AOPC-results} indicate that: 1) Grad-SAM outperforms the other methods, hence is capable of identifying the words in the input sequence that contribute the most to the (correct) model prediction. 
For example, for a BERT model that was finetuned on the SST dataset, we observe that by masking the top 20\% words proposed by {\tt [CLS]} Att, the accuracy drops to 16.5 points, whereas in the case of Grad-SAM, the accuracy drops to 19.5 points.
2) BERT is sensitive to the context;
omitting important words hinder the semantics in the sentence and significantly affects the
model’s predictions.
Overall, this AOPC evaluation provides another evidence that Grad-SAM is a state-of-the-art machinery that generates faithful explanations.

\subsection{Qualitative Results}
\label{sec:qualitative_results}
In this section, we provide qualitative examples produced by our Grad-SAM method and the {\tt {[CLS]}} Att methods from \cite{jain-etal-2020-learning}.  We follow the same procedure described in Sec.~\ref{subsec:setup_eval}: Namely, we applied both Grad-SAM and {\tt {[CLS]}} Att to rank the tokens according to their importance and considered the top $k=20\%$ tokens in the list produced by each method. Finally, we masked all the tokens in the sentences besides the top $k=20\%$ selected tokens, fed the masked sentence to BERT, and performed the prediction.

From the AGNews test set, we randomly picked $4$ examples associated with several ground truth labels. From the SST test set, we randomly picked $3$ positive and $3$ negative sentences (according to the ground truth labels). For all examples, the original prediction made by BERT (without masking) is correct (matches the ground truth label).
\begin{table}[t]
    \centering
    \scriptsize
    \setlength\tabcolsep{0.5pt}
    \begin{tabular}{l|c|c|c|c|c}
        \toprule
        Document / Sentence  & \makecell{Tokens \\highlighted \\by Grad-SAM}  & \makecell{ Tokens\\ highlighted by\\ {\tt [CLS]} Att}
        & \makecell{Original \\  Prediction}  & \makecell{Prediction \\ (Grad-SAM \\ Masking)} & \makecell{Prediction \\ ({\tt [CLS]} Att \\ Masking)} \\
        \midrule
      \multicolumn{6}{|c|}{\textbf{AGNews}} \\
        \midrule
       
        \makecell[l]{The federal agency \\ that protects
        private \\ sector pension plans \\ announced 
        yesterday \\
        that the maximum \\ annual benefit for \\ plans taken over in\\  2005 will be \$45,614 \\
        for workers who wait \\ 
        until age 65 to retire.}
        & \makecell{
        federal,\\ pension,\\ yesterday,\\ plans,\\ 2005, \\ \$45,614,\\ workers}
        & \makecell{
        annual,\\ workers,\\ wait,\\ until,\\ age, 65,\\ retire }
        &  Business 
        &  Business 
        &  World \\
        \midrule
        
        \makecell[l]{
        South Korea's key allies\\
        play down a shock admission \\
        its scientists experimented \\
        to enrich uranium. }
        & \makecell{Korea,\\ allies,\\ uranium }
        &  \makecell{s,\\ scientists,\\ enrich,\\ uranium }
        &  World 
        &  World 
        &  Sci/Tech \\
        \midrule
        
        \makecell[l]{
        OTTAWA -- A local firm that \\says it can help shrink \\ backup times at large data \\ centers
        is growing its business \\thanks
        to an alliance with \\
        Sun Microsystems Inc.}
        & \makecell{
        OTTAWA,\\
        --, local,\\
        firm,\\centers,\\
        business }
        &  \makecell{it,\\
        backup,\\ data,\\
        centers,\\
        business,\\
        Microsystems}
        &  Business 
        &  Business 
        &  Sci/Tech \\
        \midrule

        
        \makecell[l]{
        Tokyo share prices fell \\steeply Friday, led by \\ technology stocks after \\ a
        disappointing report \\ from US chip
        giant Intel. \\ The US dollar was up \\ against the Japanese yen.}
        & \makecell{Tokyo,\\ share,\\ prices,\\ Friday,\\ stocks,\\ US}
        &  \makecell{prices,\\ steeply,\\ Friday,\\ chip,\\ Intel,\\ yen}
        &  Business
        &  Business
        &  Sci/Tech \\

        \midrule
        \multicolumn{6}{|c|}{\textbf{SST}} \\
        \midrule

        \makecell[l]{A great idea becomes a \\ not great movie}
          &  not, great
          &  becomes, great 
          & Negative
          & Negative
          & Positive \\
        \midrule
        \makecell[l]{
        Flashy pretentious and as \\ impenetrable
        as morvern's thick \\ working class  scottish accent} 
          &  \makecell{flashy,\\ pretentious,\\ impenetrable}
          &  \makecell{flashy, and,\\ impenetrable} 
          & Negative
          & Negative
          & Positive \\
         \midrule
        \makecell[l]{A strong first quarter slightly \\ less so second quarter and \\ average second half}
          & \makecell[c]{strong, less, \\ average}
          & \makecell[c]{strong, and, \\ average}
          & Negative
          & Negative
          & Positive \\
          \midrule
        \makecell[l]{
        An impressive if flawed effort \\that
        indicates real talent} 
          & \makecell[c]{ impressive, \\flawed }
          & \makecell[c]{ an, flawed}
          & Positive
          & Positive
          & Negative \\
        \midrule
        \makecell[l]{
        This road movie gives you \\ emotional
        whiplash and \\you'll be glad you        went \\ along for the ride}
          &  \makecell{gives, \\ emotional,\\ whiplash, \\ glad}
          &  \makecell{this, \\ emotional,\\ whiplash, \\ and} 
          & Positive
          & Positive
          & Negative \\
        \midrule
        \makecell[l]{It never fails to engage us}
          &  never, fails
          &  it, never 
          & Positive
          & Positive
          & Negative \\
        
        \bottomrule
    \end{tabular}
    \caption{Top $k=20\%$ ranked tokens for the AGNews dataset followed by SST dataset. The tokens are ordered according to their scores in a descending order. Original Predicted stands for the prediction made by BERT on the original input (without masking). The last two columns present the prediction made by BERT after applying the masking produced by Grad-SAM and {\tt [CLS]} Att~\cite{jain-etal-2020-learning}.} 
    \label{table:qualitative-results}
\end{table}

Table ~\ref{table:qualitative-results} presents the results for both datasets. For AGNews, we observe that Grad-SAM based masking (fifth column) does not lead to a change in the model's predictions, while {\tt {[CLS]}} Att based masking (last column) does change the model's prediction, and to an incorrect one (recall that the original prediction made by the model matches the ground truth label). Finally, Grad-SAM identifies tokens that better explain the prediction made by BERT.

\section{Conclusion}
This work joins a growing effort to better interpreting deep contextualized language models.
To this end, we present \xx, a novel gradient-based method for explaining predictions made by a finetuned BERT model. Extensive evaluations show that \xx\xspace outperforms other state-of-the-art methods across various datasets, tasks, and evaluation metrics.

\bibliography{anthology,custom,references}
\bibliographystyle{ACM-Reference-Format}

\end{document}